\documentclass{article}

\usepackage{graphicx} 
\usepackage{subfigure} 

\usepackage{natbib}

\usepackage{algorithm}
\usepackage{algorithmic}

\usepackage{hyperref}

\usepackage{tikz}
\usetikzlibrary{shapes,snakes}

\usepackage{url,xspace}
\usepackage{amssymb, amsbsy, amsmath}

\renewcommand{\Re}{\mathbb{R}}
\newcommand{\dissim}{d}
\newcommand{\bbh}{\boldsymbol{h}}
\newcommand{\bbl}{\boldsymbol{\ell}}
\newcommand{\bbt}{\boldsymbol{t}}
\newcommand{\bbhp}{\boldsymbol{h'}}
\newcommand{\bbtp}{\boldsymbol{t'}}
\newcommand{\bbL}{\boldsymbol{L}}
\newcommand{\bx}{\boldsymbol{x}}
\newcommand{\by}{\boldsymbol{y}}
\newcommand{\sents}{N}
\newcommand{\norm}[1]{\parallel\! #1\! \parallel}

\newcommand{\freebase}{Freebase\xspace}
\newcommand{\wordnet}{WordNet\xspace}
\newcommand{\se}{{\sf SE}\xspace}
\newcommand{\sme}{{\sf SME}\xspace}
\newcommand{\smel}{{\sf SME}(linear)\xspace}
\newcommand{\smeb}{{\sf SME}(bilinear)\xspace}
\newcommand{\unstr}{{\sf Unstructured}\xspace}
\newcommand{\us}{{\sf Our Approach}\xspace}

\newcommand{\su}{{\it head}\xspace}
\newcommand{\ve}{{\it label}\xspace}
\newcommand{\ob}{{\it tail}\xspace}

\usepackage[accepted]{icml2013}

\icmltitlerunning{Irreflexive and Hierarchical Relations as Translations}

\begin{document} 

\twocolumn[
\icmltitle{Irreflexive and Hierarchical Relations as Translations}

\icmlauthor{Antoine Bordes}{antoine.bordes@utc.fr}
\icmlauthor{Nicolas Usunier}{nicolas.usunier@utc.fr~}
\icmlauthor{Alberto Garc\'ia-Dur\'an}{alberto.garcia-duran@utc.fr~}
\icmladdress{Heudiasyc UMR CNRS 7253,
            Universit\'e de Technologie de Compi\`egne, Compi\`egne, France}
\icmlauthor{Jason Weston}{jweston@google.com~}
\icmlauthor{Oksana Yakhnenko}{oksana@google.com~}
\icmladdress{Google,
            111 8th avenue, New York, NY, USA}

\icmlkeywords{}

\vskip 0.3in
]

\begin{abstract} 
We consider the problem of embedding entities and relations of
knowledge bases in low-dimensional vector spaces.
Unlike most existing approaches, which  are primarily efficient for modeling
equivalence relations, our approach is designed to explicitly model irreflexive relations,
such as hierarchies, by interpreting them as translations operating on the low-dimensional 
embeddings of the entities.
Preliminary experiments show that,
despite its simplicity and a smaller number of
parameters than previous approaches, our approach achieves
state-of-the-art performance according to standard evaluation
protocols on data from \wordnet and \freebase.

\end{abstract} 

\section{Intoduction}

Multi-relational data, which refers to directed graphs whose nodes correspond to {\it entities} and {\it edges} represent
 relations that link these entities, plays a pivotal role in many areas such as
recommender systems, the Semantic Web, or computational biology. 
Relations are modeled as triplets of the form (\su, \ve, \ob), where \ve indicates the type of link between the
entities \su and \ob. Relations are thus of several types and can exhibit various properties (symmetry, transitivity, irreflexivity, etc.).
%
%
Such graphs are popular tools for encoding data via knowledge bases (KBs),
semantic networks or any kind of database following the Resource Description
Framework format.
Hence, they are widely used in the Semantic Web (e.g. \freebase\footnote{\url{freebase.com}}
 or Google Knowledge Graph 
 but also for knowledge
management in bioinformatics (e.g. GeneOntology\footnote{\url{geneontology.org}})
or natural language processing
(e.g. \wordnet\footnote{\url{wordnet.princeton.edu}.}).

Despite their appealing ability for representing complex data, multi-relational databases
remain complicated to manipulate because of the heterogeneity of the relations (frequencies, connectivity), 
their inherent noise (collaborative or semi-automatic creation) and their very large dimension (up to millions of entities and thousands of relation types).

In this paper, we introduce a distributed model, which learns to embed such data in a  vector space, 
where entities are modeled as low-dimensional embeddings.
Many existing approaches (e.g. from \citet{Sutskever:2009,Nickel:2011}) 
interpret relations as linear transformations
of these embeddings: when $(h,\ell,t)$ holds, then the embeddings of
\su $h$ and \ob $t$ should be close (in the embedding space) after
transformation by a linear operator that depends on the \ve
$\ell$. With such an interpretation, the model implies that the
relation is reflexive since the embedding of $h$ will always be its
nearest neighbor, and because of the triangle inequality, the model
will, to some extent, imply some form of transitivity of the relation.

While this interpretation is fine for equivalence relations (such as
\wordnet's {\tt \_similar\_to}), it is inadequate for irreflexive
relations that represent hierarchies, such as \wordnet's {\tt
  \_hypernym} or \freebase's type hierarchy. Indeed, taking the
simplest example of entities organized in a tree with two relations,
``sibling'' and ``parent'', the embeddings of siblings should be close
to each other (since it essentially is an equivalence relation), but
enforcing the constraint that parent nodes should be close to their
child nodes will lead the embedding of the whole tree to collapse
to a small region of the space where the siblings and parent of a
given node are impossible to distinguish.

Since hierarchical and irreflexive relations are extremely common in
KBs, we propose a simple model to efficiently represent them, by
interpreting {\it relations as translations in the embedding space}:
if $(h,\ell,t)$ holds, then the embedding of $t$ should be close to
the embedding of $h$ plus some vector that depends on $\ell$.
This approach is motivated by the natural representation of trees
(i.e. embeddings of the nodes in dimension 2): while siblings are
close to each other and nodes at a given height are organized on the
$x$-axis, the parent-child relation corresponds to a translation on
the $y$-axis. Another, secondary, motivation comes from the recent
work of \citet{DBLP:journals/corr/abs-1301-3781}, in which the authors
learn word embeddings from free text, and some one-to-one relations
between entities of different types, such as the relation ``capital
of'' between countries and cities, are (coincidentally rather than willingly)
represented by the model as translations in the embedding space. Our
approach may then be used in the context of learning word embeddings
in the future to reinforce this kind a structure of the embedding
space.


%
%
%
Apart from the main line of algorithms to learn embeddings of KBs, a
number of recent approaches deal with the asymmetry of the relations at
the expense of an explosion of model parameters. We present an
empirical evaluation on data dumps of \wordnet and \freebase, in which
our model achieves strong results compared to such algorithms, with
much fewer parameters and even lower dimensional embeddings.



In the remainder of the paper,
%
we describe some of the related work in Section~\ref{sec:rwork}. We then
describe our model in Section~\ref{sec:model}, and discuss its
connections with related methods. 
We report preliminary experimental results on
\wordnet and \freebase in
Section~\ref{sec:exp}. We finally sketch some future
work directions in Section~\ref{sec:concl}.

\section{Related work}\label{sec:rwork}

Most previous methods designed to model relations in multi-relational
data rely on latent representations or embeddings.
The simplest form of latent attribute that can be associated to an entity is a latent class. Several clustering approaches 
have been proposed. \citet{Kemp:2006} considered a non-parametric Bayesian extension of the \emph{stochastic blockmodel} allowing to automatically infer the number of latent clusters; \citet{Kok:2007} introduced clustering in Markov-Logic networks; \citet{Sutskever:2009} used a non-parametric Bayesian clustering of entities embedding in a \emph{collective matrix factorization} formulation. All these models cluster not only entities but relation labels as well.

These methods can provide interpretations and analysis of the data but are
slow and do not scale to large databases, due to the high cost of inference.
In terms of scalability, models based on tensor factorization (like those from \citep{Harshman:1994} or \cite{Nickel:2011}) have shown to be efficient. However, they have been outperformed by energy-based models \cite{bordesAAAI11,jenatton2012latent,bordesMLJ2013,chen2013}. These methods represent entities as low-dimensional embeddings and relations as linear or bilinear operators on them and are trained via an online process, which allows them to scale well to large numbers of entities and relation types.
In Section~\ref{sec:exp}, we compare our new approach to \se \cite{bordesAAAI11} and \sme \cite{bordesMLJ2013}.

\section{Translation-based model}\label{sec:model}

We now describe our model and discuss its relationship to existing
approaches.

\subsection{Our model}

Given a training set $S$ of labeled arcs $(h,\ell,t)$,
our goal is to learn vector embeddings for all values of $h$, $\ell$
and $t$. We assume all nodes and labels appear at least once in
the training set. The embeddings take values in $\Re^k$ ($k$ is a
model hyperparameter) and are denoted with the same letter, in
boldface characters. The basic idea behind our model is that the
functional relation induced by the $\ell$-labeled arcs corresponds to
a translation of the embeddings, i.e. we want that $\bbh+\bbl \approx
\bbt$ when $(h,\ell,t)$ holds, while $\bbh+\bbl$ should be far away
from $\bbt$ otherwise.

To learn such embeddings, we minimize the following margin-based
ranking criterion over the training set:
\begin{equation}
\label{eq:objectiveFunc}
\sum_{(h,\ell,t) \in S} \sum_{(h', \ell, t')\in S'_{(h,\ell,t)}} \hspace{-0.6cm}\big[\gamma+\dissim(\bbh+\bbl, \bbt)-\dissim(\bbhp+\bbl, \bbtp)\big]_+
\end{equation}
where $[x]_+$ denotes the positive part of $x$, $\gamma>0$ is a margin
hyperparameter, $\dissim(\bx, \by)$ is some dissimilarity function on
$\Re^k$, e.g. the euclidian distance or the squared euclidian
distance, and
\begin{equation}
\label{eq:negativeExs}
S'_{(h,\ell,t)}=\big\{(t', \ell,t) | h'\in \sents\big\}\cup \big\{(h, \ell, t')|t'\in \sents \big\}\,.
\end{equation}
The set of ``negative'' examples we sample according to Equation
\ref{eq:negativeExs} is basically the training (``positive'') triple
with either the head or tail replaced by a random entity (but not both
at the same time). The loss function \eqref{eq:objectiveFunc} favors
low values of dissimilarity between head+label and tail for positive
triplets, and large values for negative triplets, and is thus a natural
implementation of the intended criterion.

The minimization is carried out by stochastic
gradient descent, over the possible $\bbh, \bbl$ and $\bbt$, with the
additional constraints that the $L_2$-norm of the embeddings of the
entities is $1$ (no regularization or norm constraints
are given to the label embeddings $\bbl$).

\subsection{Relationship to previous approaches}\label{sec:pap}

Section~\ref{sec:rwork} described a large body of work on embedding KBs. 
We detail here the relationships between our model and those of \citet{bordesAAAI11} (Structured
Embeddings or \se) and \citet{chen2013}.

\se \citep{bordesAAAI11} embeds nodes into $\Re^k$, and
labels into two matrices $\bbL_1 \in \Re^{k \times k}$ and $\bbL_2 \in
\Re^{k \times k}$ such that $\dissim(\bbL_1 \bbh, \bbL_2\bbt)$ is
small for positive triplets $(h,\ell, t)$ (and large otherwise). The
basic idea is that when two nodes belong to the same edge, their
embeddings should be close to each other in some subspace that depends
on the label. This basic idea would imply $\bbL_1=\bbL_2$, and using
two different projection matrices for the head and for the tail is
intended to account for the possible asymmetry of relation
$\ell$. When the dissimilarity function takes the form of
$\dissim(\bx, \by) = g(\bx-\by)$ for some $g:\Re^k\rightarrow \Re$
(e.g. $g$ is a norm), then the model of \se with an
embedding of size $k+1$ is strictly more expressive than our model
with an embedding of size k, since linear operators in dimension $k+1$
can reproduce affine transformations in a subspace of dimension $k$
(by constraining the $k+1$st dimension of all node embeddings to be
equal to $1$). \se, with $\bbL_2$ as the
identity matrix and $\bbL_1$ taken so as to reproduce a translation is
then equivalent to our model. Despite the lower expressiveness of our
model, we still reach better performance than this model in our
experiments because (1) our model is a more direct
way to represent the true properties of the relations, and (2)
regularization, and more generally any form of capacity control, is
difficult in embedding models ; greater expressiveness may then be more
synonymous to overfitting than to better performance.

Another related model is the Neural Tensor Model of
\citet{chen2013}. A special case of that model (which would actually
boil down to a ``Neural Matrix Model'') corresponds to learn scores
$s(h,\ell,t)$ (higher scores for positive triplets) of the form:
\begin{equation}
\label{eq:chenmodel}
s(h, \ell, t) = \bbh^T \bbL \bbt + \bbl_1^T\bbh +\bbl_2^T\bbt
\end{equation}
where $\bbL \in \Re^{k \times k}$, $\bbL_1\in\Re^k$ and
$\bbL_2\in\Re^k$, all of them depending on $\ell$.

If we consider our model with the squared 
distance as dissimilarity function, we have:
\begin{equation*}
d(\bbh+\bbl, \bbt) = \norm{\bbh}^2+\norm{\bbl}^2+\norm{\bbt}^2 -2 \big(\bbh^T\bbt+\bbl^T(\bbt-\bbh)\big)\, . 
\end{equation*}
Considering our norm constraints ($\norm{\bbh}^2=\norm{\bbt}^2=1$) and
the ranking criterion \eqref{eq:objectiveFunc}, in which
$\norm{\bbl}^2$ does not play any role in comparing positive and
negatives triplets, our model thus corresponds to the scoring triplets
according to $\bbh^T\bbt+\bbl^T(\bbt-\bbh)$, and thus corresponds to
\citet{chen2013}'s model (Equation \eqref{eq:chenmodel}) where $\bbL$ is
the identity matrix, and $\bbl=\bbl_1=-\bbl_2$. We could not run
experiments with that model, but once again our model has much fewer
parameters: this should ease the training and prevent overfitting, and hence compensate for a lower expressiveness.

\begin{table}[t]
\caption{Statistics of the data sets used in this paper.}
\label{tab:data}
\vskip 0.15in
\begin{center}
\begin{small}
\begin{sc}
\begin{tabular}{l|rr}
\hline
\abovespace\belowspace
Data set & \wordnet & \freebase \\
\hline
\abovespace
Entities    & 40,943& 14,951\\
Rel. types & 18& 1,345\\
Train. ex.    & 141,442& 483,142\\
Valid ex.    & 5,000& 50,000\\
Test ex.     & 5,000& 59,071\\
\hline
\end{tabular}
\end{sc}
\end{small}
\end{center}
\vskip -0.1in
\end{table}

\begin{table*}[t]
\caption{Some example predictions on the \freebase test set using our approach. {\bf Bold} indicates the test triple's true tail and {\it italics} other true tails present in the training set. Actual Freebase identifiers  have been replaced by readable strings. }
\label{tab:fb_ex}
\vskip 0.15in
\begin{center}
\begin{scriptsize}
\begin{tabular}{l|c}
\hline
\abovespace\belowspace
\footnotesize{\sc Input (Head and Label)} &  \footnotesize{\sc Predicted Tails} \\
\hline
J. K. Rowling {\tt  influenced by} &  {\em G. K. Chesterton}, J. R. R. Tolkien, {\em C. S. Lewis}, {\bf Lloyd Alexander},\\
&  Terry Pratchett, Roald Dahl, Jorge Luis Borges, {\em Stephen King}, Ian Fleming \\
\hline
Anthony LaPaglia {\tt  performed in} & {\em Lantana}, {\em Summer of Sam}, {\em Happy Feet}, 
{\em The House of Mirth}, \\ 
& Unfaithful, {\bf Legend of the Guardians}, Naked Lunch, X-Men, The Namesake\\
\hline
Camden County  {\tt  adjoins} & {\bf Burlington County}, {\em Atlantic County}, {\em Gloucester County},
Union County, \\
& Essex County, New Jersey, Passaic County, Ocean County, Bucks County \\
\hline
The 40-Year-Old Virgin {\tt  nominated for} & 
   {\em MTV Movie Award for Best Comedic Performance,} \\
& {\em BFCA Critics' Choice Award for Best Comedy,} \\
& {\em MTV Movie Award for Best On-Screen Duo,} \\
&  MTV Movie Award for Best Breakthrough Performance, \\
& {\bf MTV Movie Award for Best Movie}, MTV Movie Award for Best Kiss, \\
& D. F. Zanuck Producer of the Year Award in Theatrical Motion Pictures, \\
& Screen Actors Guild Award for Best Actor - Motion Picture \\
\hline
 David Foster {\tt  has the genre} & 
{\em Pop music}, {\em Pop rock}, Adult contemporary music, Dance music, \\
& {\bf Contemporary R\&B}, Soft rock, Rhythm and blues, Easy listening \\
\hline
Costa Rica football team {\tt  has position} & 
{\em Forward}, {\em Defender}, {\em Midfielder}, {\bf Goalkeepers}, \\
& Pitchers, Infielder, Outfielder, Center, Defenseman \\
\hline
Lil Wayne {\tt  born in} & {\bf New Orleans}, Atlanta, Austin, St. Louis,\\
&  Toronto, New York City, Wellington, Dallas, Puerto Rico  \\
\hline
WALL-E {\tt  has the genre} & Animations, Computer Animation, {\em Comedy film}, \\
& {\em Adventure film}, {\em Science Fiction}, {\bf Fantasy}, Stop motion, {\em Satire}, Drama \\
\hline
Richard Crenna {\tt  has cause of death} & {\em Pancreatic cancer},  {\bf Cardiovascular disease}, Meningitis, Cancer, \\
& Prostate cancers, Stroke, Liver tumour, Brain tumor, Multiple myeloma \\
\hline
\end{tabular}
\end{scriptsize}
\end{center}
\vskip -0.1in
\end{table*}

\section{Experiments} \label{sec:exp}

Our approach is evaluated against the methods \se and \sme (Semantic Matching Energy) from
\citep{bordesAAAI11,bordesMLJ2013} on two data sets and using the same ranking setting for evaluation.

%
We measure the mean and median predicted ranks and the top-10, computed with the following
procedure.
For each test triplet, the head is removed and replaced by each of the
entities of the dictionary in turn. Energies (i.e. dissimilarities) of
those corrupted triplets are computed by the model and sorted by
ascending order and the rank of the correct entity is stored. This
whole procedure is also repeated when removing the tail instead or the head.  We
report the mean and median of those predicted ranks and the top-10,
which is the proportion of correct entities in the top 10 ranks.

\subsection{Data}

We used data from two KBs; their statistics are given in Table~\ref{tab:data}.

\paragraph{\wordnet}

This knowledge base is designed to produce an intuitively usable dictionary and thesaurus,
and support automatic text analysis. Its entities (termed
{\it synsets}) correspond to word senses, and relation types define lexical relations between
them.
We considered the data version used in \citep{bordesMLJ2013}. 
Examples of triplets are ({\it \_score\_NN\_1}, {\it \_hypernym}, {\it
  \_evaluation\_NN\_1}) or ({\it \_score\_NN\_2}, {\it \_has\_part}, {\it
  \_musical\_notation\_NN\_1}).\footnote{\wordnet is composed of senses, its entities are termed by the
concatenation of a word, its part-of-speech tag and a digit indicating which sense it refers to i.e. {\it
  \_score\_NN\_1} encodes the first meaning of the noun ``score''.}

\paragraph{\freebase}

Freebase is a huge and growing database of
general facts; there are currently around 1.2 billion triplets. To make
a small data set to experiment on we selected the subset of
entities that are also present in the Wikilinks database\footnote{\url{code.google.com/p/wiki-links}}
 and that also have at least 100 mentions in Freebase (for both
entities and relations).  We also removed negative relations like
'!/people/person/nationality' which just reverses the head and tail
compared to the relation '/people/person/nationality'.  This resulted in 592,213 triplets
with 14,951 entities and 1,345 relations which were randomized and split as shown in
Table \ref{tab:data}.

\subsection{Implementation}

We implemented our model using the SME library\footnote{\url{https://github.com/glorotxa/SME}}, which already 
proposes code for \se and \sme. The dissimilarity measure $\dissim$ was set to the $L_1$ distance, mostly because it led to a faster training.

For this preliminary set of experiments, we did not perform an extensive search 
for hyperparameters. 
For experiments of our method on \wordnet, we fixed the learning rate for the stochastic gradient descent to $0.01$, the dimension $k$ of the embeddings to $20$ and chosen the margin $\gamma$ among $\{1,2,10\}$ with the validation set (optimal value was $2$).
We report results for \se and \sme extracted from \citep{bordesMLJ2013} where those models have been trained using a much more thorough hyperparameter search.
For experiments on \freebase, we ran all experiments using the SME library with fixed values for the learning rate ($=0.01$), $k$ ($=50$) and $\gamma$ ($=1$).
For both datasets, the training time was limited to at most $1,000$ epochs over the training set. The best model was selected using the mean predicted rank on the validation set.

\subsection{Results}

\begin{table}[t]
\caption{Link prediction results on \wordnet.}
\label{tab:wn}
\vskip 0.15in
\begin{center}
\begin{small}
\begin{sc}
\begin{tabular}{l|cc|c}
\hline
\abovespace
Method & \multicolumn{2}{c|}{Rank} & Top-10 \\
\belowspace
& Mean  & Med.  &  \\
\hline
\abovespace
\unstr    	& 317 		& 26 	& 35.1\%		\\
\se 		& 1,011		& \bf 3 & 68.5\%		\\
\smel    	& 559		& 5 	& 65.1\%	\\
\smeb    	& 526 		& 8 	& 54.7\%		\\
\us     	& \bf 263 	& \bf 4 & \bf 75.4\%	\\
\hline
\end{tabular}
\end{sc}
\end{small}
\end{center}
\vskip -0.2in
\end{table}

\begin{table}[t]
\caption{Link prediction results on \freebase.}
\label{tab:fb}
\vskip 0.15in
\begin{center}
\begin{small}
\begin{sc}
\begin{tabular}{l|cc|c}
\hline
\abovespace
Method & \multicolumn{2}{c|}{Rank} & Top-10 \\
\belowspace
& Mean  & Med.  &  \\

\hline
\abovespace
\unstr    	& 1097 		& 404 	& 4.5\%		\\
\se 		& 272		& 38 	& 28.8\%		\\
\smel    	& 274		& 34 	& 30.7\%		\\
\smeb    	& 284 		& 35 	& 31.3\%		\\
\us   	& \bf 243 	& \bf 25 & \bf 34.9\%	\\
\hline
\end{tabular}
\end{sc}
\end{small}
\end{center}
\vskip -0.1in
\end{table}

Tables~\ref{tab:wn} and~\ref{tab:fb} displays the results on both data sets for our method, 
compared to \se, to two versions of \sme and to \unstr, a simple model which only uses the 
dot-product between $\bbh$ and $\bbt$ as dissimilarity measure for a triplet $(h,\ell,t)$, with no influence of $\ell$.
Table~\ref{tab:fb_ex} gives examples of nearest link prediction
results of our approach on the \freebase test set.


Our method greatly outperforms all counterparts on all metrics, with particularly good results for the top-10 metric. 
We believe that such remarkable performance is due to
an appropriate design of the model according to the data, but also to its relative simplicity.
Hence, even if the problem is non-convex, it can be optimized efficiently with stochastic gradient. We showed in Section~\ref{sec:pap} that \se is more expressive than our proposal. However, its complexity makes it quite hard to train as shown in the results of tables~\ref{tab:wn} and~\ref{tab:fb}.

Table~\ref{tab:fb_ex} illustrates the capabilities of our model. Given a head and a label, the top predicted tails (and the true one) are depicted. The examples come from the \freebase test set. Even if the good answer is not always top-ranked, the predictions reflect common-sense.

\section{Conclusion and future work} \label{sec:concl}

We proposed a new approach to learn embeddings of KBs, focusing on the
minimal parametrization of the model to accurately represent
hierarchical and irreflexive relations. This short paper is
essentially intended to be a proof-of-concept that translations are
adequate to model such relations in a multi-relational setting. 
It can be improved and better validated in several ways.
For the experimental evaluation, this paper is the first one to
present link prediction on this dump of \freebase. More benchmarking 
is needed, such as the comparison with models of \citet{chen2013} and
\citet{jenatton2012latent}. We also intend to consider learning
translations of word embedding, either from free text as in
\cite{DBLP:journals/corr/abs-1301-3781} or from $(subject, verb,
object)$ triplets as in \cite{bordesAAAI11}.

Finally, regarding modeling relations, equivalence
relations in our approach are represented by a $\boldsymbol 0$
translation vector, and thus enforces all members of an equivalence
class to be close to each other in the embedding space (whatever the
relation). Some additional degrees of freedom may be given by adding a
projection matrix to each relation, so that equivalence relations only
enforce entities to be close to each other in some subspace of the
embedding space. However, this would increase the number of parameters, and 
we believe that regularization and optimization
techniques should be further studied to achieve optimal performance.




\section*{Acknowledgments} 
We thank Thomas Strohmann and Kevin Murphy for useful discussions.
This work was supported by the French ANR (EVEREST-12-JS02-005-01).
{{\small
\bibliographystyle{icml2013}
\bibliography{sme_mlj.bib}
}}

\end{document}